\documentclass[review]{elsarticle}

\usepackage{lineno,hyperref}
\usepackage{subcaption}
\usepackage{overpic}
\modulolinenumbers[5]

\begin{document}

\begin{frontmatter}

\title{Optical Flow Super-Resolution Based on Image Guidence Using Convolutional Neural Network }
%\tnotetext[mytitlenote]{Fully documented templates are available in the elsarticle package on \href{http://www.ctan.org/tex-archive/macros/latex/contrib/elsarticle}{CTAN}.}\tnoteref{mytitlenote}

%% Group authors per affiliation: \corref{cor1}
\author{Liping Zhang}

\author{Zongqing Lu\corref{Corresponding}}
%\cortext[Corresponding]{Corresponding author}

\author{Qingmin Liao\corref{}}

\cortext[Corresponding]{Corresponding author.}
\address{Graduate School at Shenzhen, Tsinghua University Shenzhen, China}

%%\author[mymainaddress,mysecondaryaddress]{Elsevier Inc}
%%\ead[url]{www.elsevier.com}

%\author[mysecondaryaddress]{Corresponding author\corref{mycorrespondingauthor}}

%%\address[mymainaddress]{1600 John F Kennedy Boulevard, Philadelphia}
%%\address[mysecondaryaddress]{360 Park Avenue South, New York}

\begin{abstract}

The convolutional neural network model for optical flow estimation usually outputs a low-resolution(LR) optical flow field.  
To obtain the corresponding full image resolution, interpolation and variational approach are the most common options, which do not effectively improve the results.
With the motivation of various convolutional neural network(CNN) structures succeeded in single-image super-resolution(SISR) task, an end-to-end convolutional neural network is proposed to reconstruct the high resolution(HR) optical flow field from initial LR optical flow with the guidence of the first frame used in optical flow estimation. Our optical flow super-resolution(OFSR) problem differs from the general SISR problem in two main aspects. Firstly, the optical flow includes less
 texture information than image so that the SISR CNN structures can't be directly used in our OFSR problem. Secondly, the initial LR optical flow data contains estimation error, while the LR image data for SISR is generally a bicubic downsampled, blurred, and noisy version of HR ground truth.
We evaluate the proposed approach on two different optical flow estimation mehods and show that it can not only obtain the full image resolution, but generate more accurate optical flow field (Accuracy   
 improve$ ~15\% $ on FlyingChairs, $ ~13\% $ on MPI Sintel) with sharper edges than the estimation result of original method.

\end{abstract}

\begin{keyword}
Optical flow\sep Super-Resolution
\end{keyword}

\end{frontmatter}

%\linenumbers

\section{Introduction}

\begin{figure}[htbp]
\centering
\includegraphics[width=.8\textwidth]{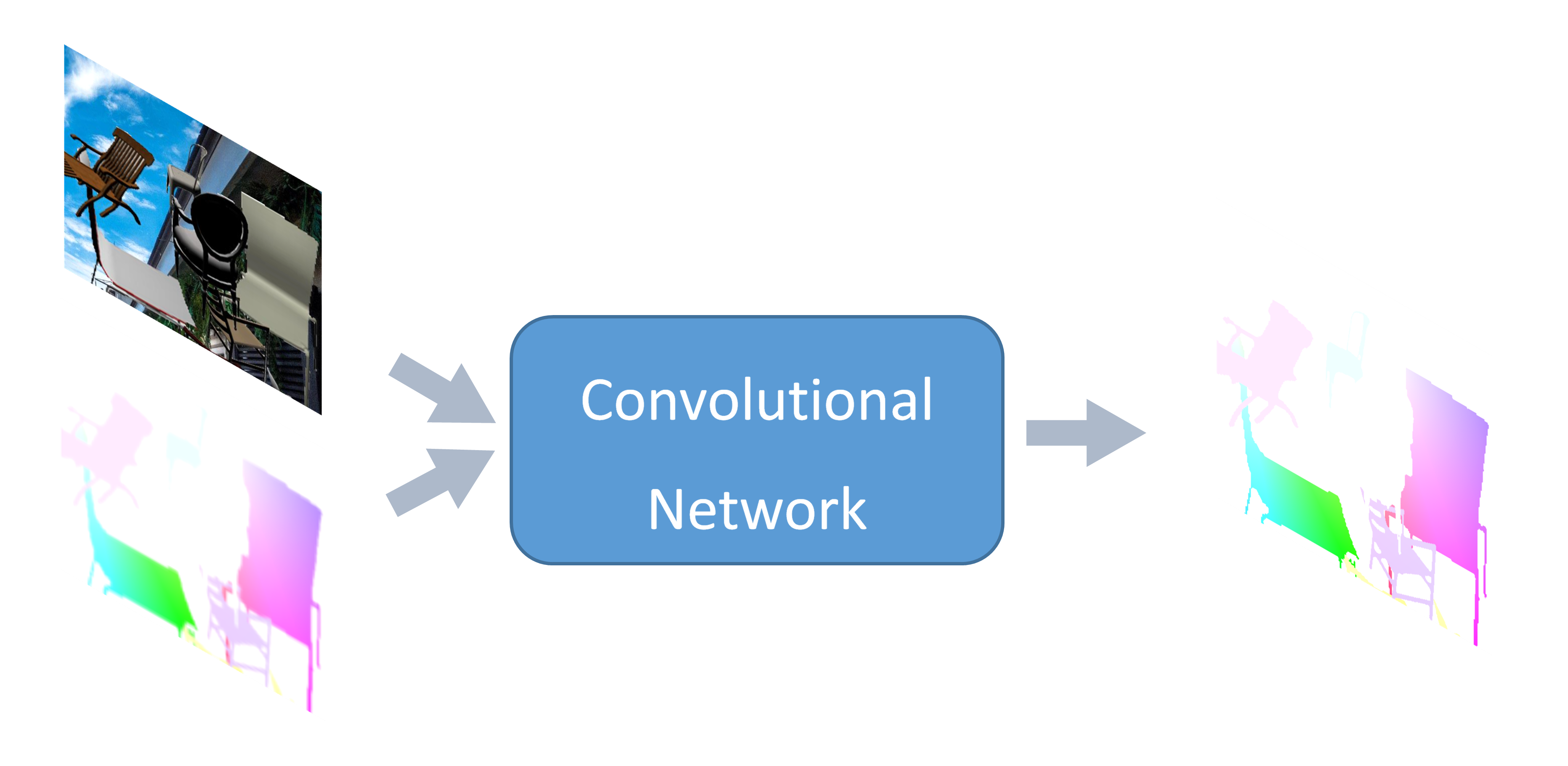}
\caption{An end to end learning convolutional neural network is proposed to realize super-resolution reconstruction of high resolution optical flow field.}
\label{network}
\end{figure}

Many optical flow estimation methods\cite{Dosovitskiy_2015_ICCV, ilg2017flownet,sun2018pwc,ranjan2017optical,zhu2017guided,bailer2015flow} use two consecutive frames images as input to obtain optical flow estimation results through their network. Bilinear upsampling\cite{Dosovitskiy_2015_ICCV, ilg2017flownet, sun2018pwc} and variational approach\cite{horn1981determining, Dosovitskiy_2015_ICCV} are often used to get full image resolution in the final stage\cite{Dosovitskiy_2015_ICCV,sun2018pwc}, which do not improve performance significantly.
In this paper, as is shown in figure \ref{network}, a CNN model is proposed to take the first frame image and the LR optical flow predicted by some existing methods as input to generate more accurate HR optical flow field.

The optical flow field is currently the most common and most effective method for describing two-dimensional motion of images\cite{barron1994performance, fortun2015optical, baker2011database}. According to the definition of image motion in Horn\cite{horn1986robot}, the optical flow field is the apparent motion of luminance mode in the image.
Specifically, for two consecutive frames of images, the optical flow field of the first frame image corresponds to a two-dimensional vector field formed by motion vectors of pixels in the first frame image\cite{beauchemin1995computation}. 
Optical flow estimation and Super-Resolution have long been the subject of much attention in computer vision tasks\cite{fortun2015optical, dong2014learning}.
In recent years, significant progress has been made in using CNN to solve the above two problems\cite{fortun2015optical, yue2016image}.

Inspired by the significant advances in convolutional network design and the successful application of CNN models to visual tasks, Dosovitskiy et al.\cite{Dosovitskiy_2015_ICCV} first proposed two end-to-end optical flow estimation CNN models, FlownetS and FlownetC.
At the end of the both network, they use two apporach to upscale the LR optical flow filed to get full image resolution. One is bilinear upsampling,
the other method they used called variational refinement, which utilize the variational approach from\cite{brox2011large} without the matching term and additional compute image boundaries with the approach in \cite{leordeanu2012efficient}. 
Sun et al.\cite{sun2018pwc} proposed PWC-Net which combines deep learning and domain knowledge\cite{black1996robust, chen2016full, xu2017accurate, brox2004high}. The PWC-Net outperformes many public methods also uses the bilinear interpolation to upsample the quarter resolution optical flow of model output.

Considering the optical flow data as a 2-channel image data, it is feasible to restore the HR optical flow field by referring to the well-performing CNN structures\cite{dong2016accelerating, kim2016accurate, kim2016deeply, shi2016deconvolution, lim2017enhanced, zhang2018residual} of solving SISR. 
But there are two major difference between our Optical Flow Super-Resolution(OFSR) problem and SISR task.
Firstly, the network input LR optical flow is calculated by some open optical flow estimation methods. According to the performance of different methods, the input LR optical flow data would contain different estimation error, which is very different from LR training data used in SISR task. The LR image is generally a bicubic downsampled, blurred, and noisy version of HR image\cite{yang2010image, park2003super, glasner2009super}.
Secondly, the optical flow contains much less textural information different from general image. Therefore, the SISR CNN model cannot be applied directly to optical flow problems.

For the OFSR problem, we assume the motion boundary of optical flow field contained in the motion object boundary of the corresponding image. The optical flow field contains sparse textural information
and its data structure is similar to the RGB image.
So the flow texture must appear on the image, the introduction of reference frame can ensure that the motion boundary information would not be destroyed, and
 can help guide the OPSR process to resume more details and clear-cut edges. Therefore, we introduce the first image as the guidance information to help improve the accuracy of the  output HR result. It means OFSR network not only needs to get HR from LR reconstruction but to suppress noise, restore details, and improve the estimation performance of the original methods. Furthermore, this method can be applied to the existing methods as a refinement approach for better performance.

The rest of this article is organized as follows.
In section 2, we present the related work of using CNN model single image SR in recent years.
Section 3 focuses on the proposed network structure, and section 4 is devoted to the implementation details and experimental results. Finally, the general conclusions are summarized in Section 5.
\section{Related Work on Image Super-Resolution}

In many deep learning-based SR algorithms, the input image is typically upsampled into the HR space\cite{dong2016image, wang2015deeply,chen2017trainable} by bicubic interpolation before being fed into the network.
This means that the image SR is performed in the HR space, and Shi et al.\cite{shi2016real} proposed a new network structure with better performance and faster speed called ESPCN.
The ESPCN extracts the feature map in the LR space, and the HR output is obtained by a valid sub-pixel convolution layer at the end of the network to enlarge the LR feature map. This method reduces the SR computational complexity very well.
Zhang et al.\cite{zhang2018residual} combined Residual block in MDSR \cite{lim2017enhanced} and Dense block in SRDenseNet\cite{tong2017image} to propose a residual dense block called RDB, following \cite{shi2016real}: they extract features in LR, use residual connection in single blocks, multiple blocks are densely connected and residual connection is used on the whole. In this way, the number of feature maps, convolution layers and blocks can be adjusted, forming a very deep CNN. Their network has a strong ability on suppressing the blurring artifacts and recovering sharper edges.

\section{Network Architectures}

The end-to-end CNN model has significant advantages in solving optical flow estimation and SISR problems, and we use this end-to-end learning method to perform optical flow SR.
In this paper, the purpose of our CNN is to reconstruct more smooth, subpixel-accurate HR optical flow field from initial LR optical flow field and first frame image.

In the calculation of optical flow estimation\cite{fleet2006optical}, for a pair of consecutive frames of images, the x-y optical flow field of a certain frame is a vector field formed by motion vectors of pixels in the corresponding frame image, therefore, the optical flow field calculated from two consecutive frames of images is structurally consistent with the first frame image.
We take the first frame as the guidance to help reconstruction process.

\begin{figure}[htbp]
\centering
\includegraphics[width=\textwidth]{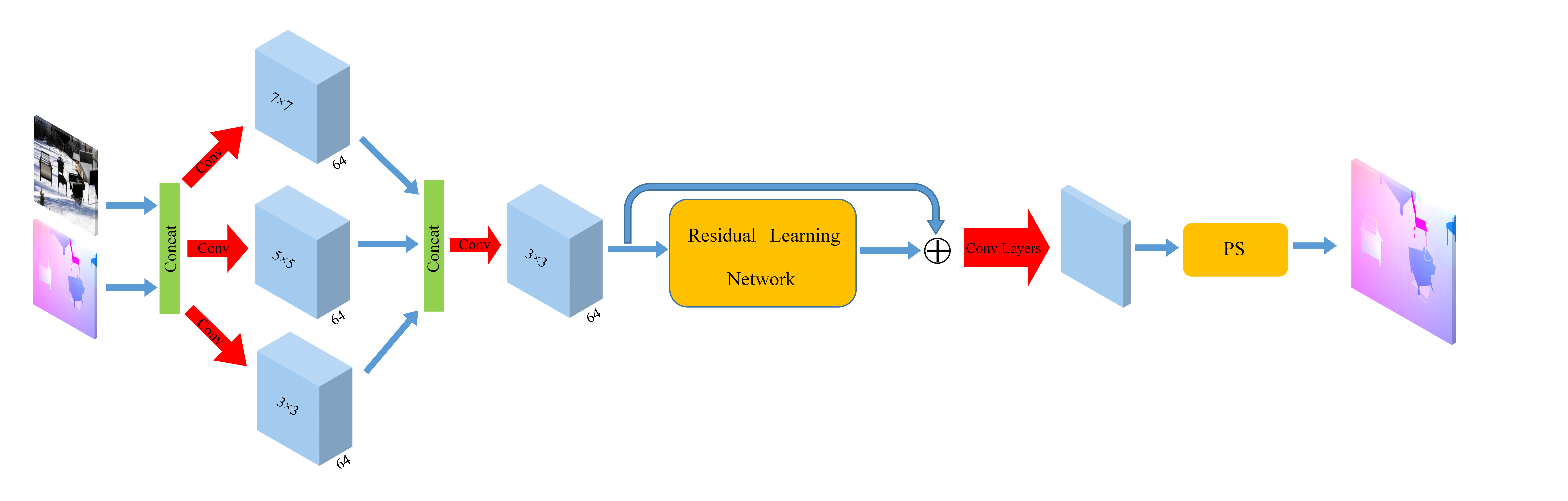}
\caption{The architecture of our proposed network for optical flow super-resolution(OFSR)}
\label{structure}
\end{figure}

As shown in the figure \ref {structure}, our OFSR network is mainly composed of three parts: feature extraction network, residual learning network and upscale network.

\begin{figure}[htbp]
\centering
\includegraphics[width=\textwidth]{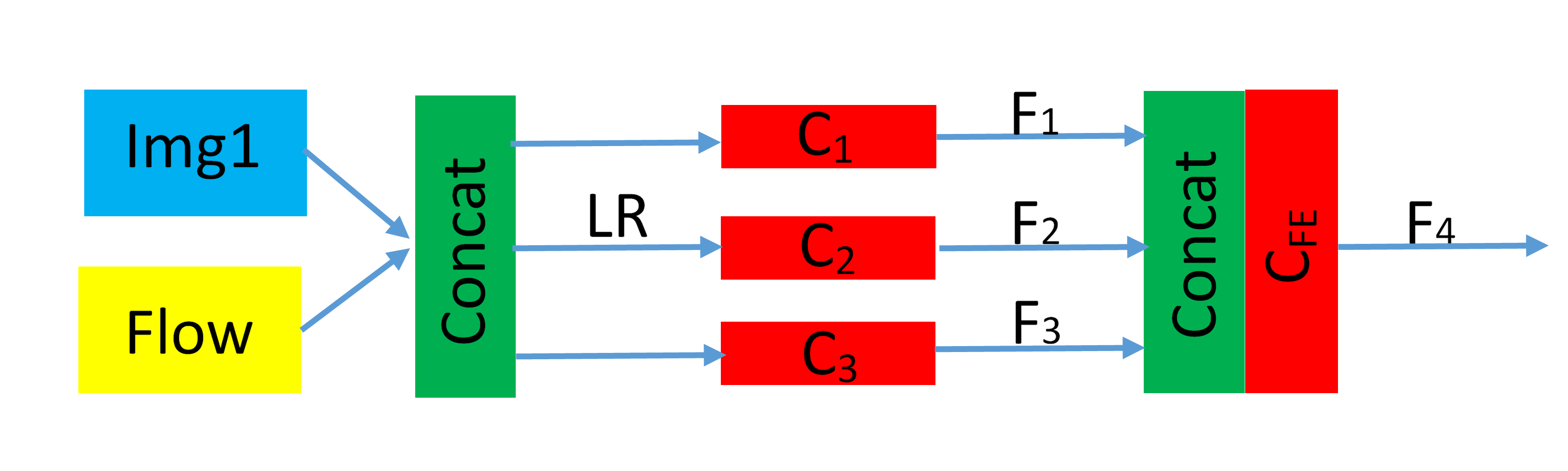}
\caption{The architecture of feature extraction network}
\label{part1}
\end{figure}

\paragraph{Feature extraction network}
Figure \ref{part1} shows the structure of our feature extraction network, the input of the network is the simple concatenation of the first frame image and LR flow.
The addition of the first frame image provides constraint and guidance of motion graphic boundary and related details for super resolution reconstruction of low resolution optical flow. Compared with SISR, the optical flow SR network is able to obtain more information for learning. In addition, we believe that the multi-scale receptive fields can further provide richer multi-scale shallow feature information for subsequent residual learning.

To get multi-scale features, as shown in the figure \ref{part1}, we first use three convolutional layers of different filter sizes to extract the shallow features separately, and then to implement feature fusion we use a layer of convolution followed by RelU to extract features from these shallow features as input to the next part of the network.

\begin{equation}
F_{i} = C_{i}(LR), i = 1, 2, 3 
\label{eq1} 
\end{equation}

Where $ C_{i} $ represents the combined operation of the convolution and activation function(RELU), and $ F_{i} $ is then used for further feature extraction.

\begin{equation}
 F_{4} = C_{FE}(F_{i}), i = 1, 2,3 
\label{eq2} 
\end{equation}

Where $ C_{FE} $ denotes the concatenation of $ F_{i} $, followed by convolution operation, and $ F_{4} $ is used as input to the residual learning network.

\paragraph{Residual learning network}
Inspired by \cite{huang2017densely}, we use the Residual dense block (RDB) in \cite{zhang2018residual} to perform residual learning on the features obtained by the feature extraction network.
In our network structure, the optical flow SR task not only needs to make full use of the shallow feature information extracted from LR optical flow field, but also needs to fully learn the effective information which can be used for reconstructing the accurate and high-resolution optical flow field from the first image. RDB has shown good performance in SISR problem, which proves that it is effective in extracting abundant feature information through dense connection.
We use this structure to extract local information repeatedly from the shallow features of LR optical flow field and the first frame image. 
The direct connection of local features extracted by multiple blocks can adaptively fuse the feature information that can help to modify and optimize the input optical flow field in the first frame image.

\begin{figure}[htbp]
\centering
\includegraphics[width=\textwidth]{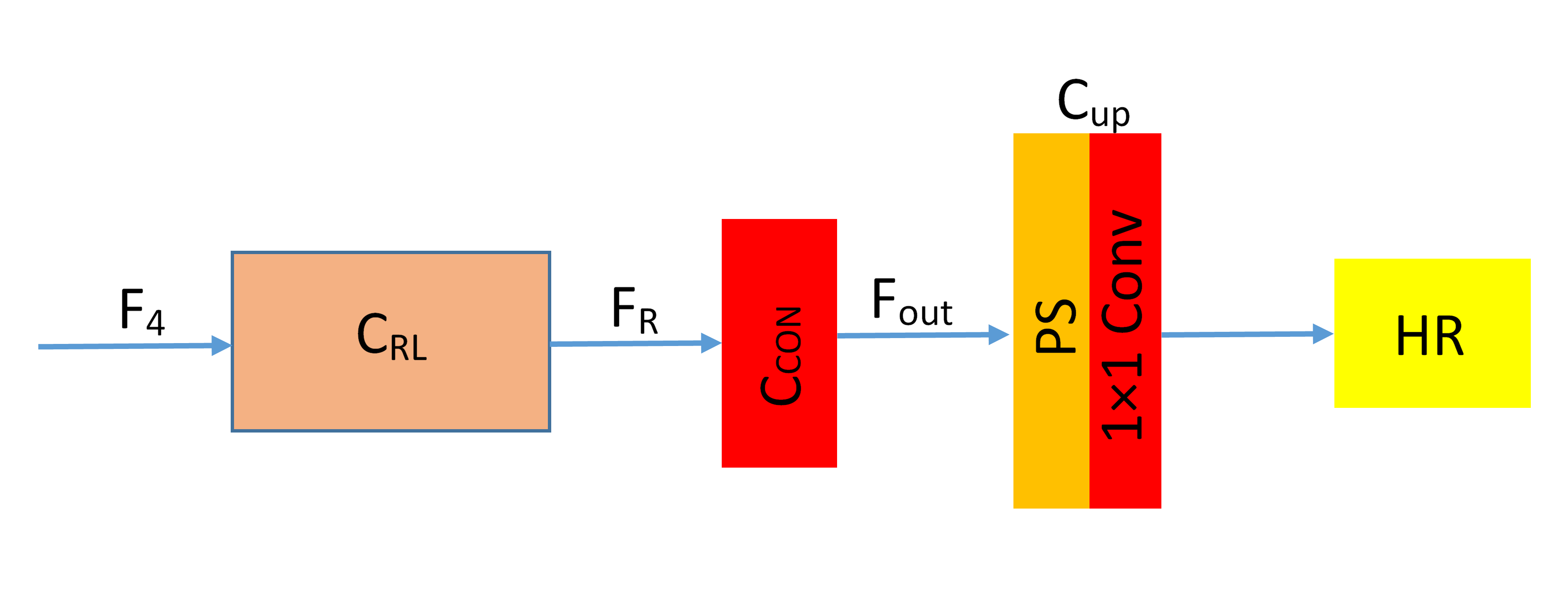}
\caption{The architecture of residual learning network and upscale network}
\label{part23}
\end{figure}
As is shown in figure \ref{part23}, $ C_{RL} $ is used to represent the operation of the residual learning network, and $F_{R}$ is the feature obtained by the residual learning in the LR space.

\begin{equation}
 F_{R} = C_{RL}(F_{4})
\label{eq3} 
\end{equation}

\paragraph{Upscale network}

In the first two parts of the network, the calculation such as feature extraction was carried out in LR space, which greatly reduces the computational complexity of the whole SR process.
Inspired by the efficient sub-pixel convolution layer proposed in \cite{shi2016real}, Our HR optical flow field output is obtained by upscaling the final feature map in the Upscale network.
Compared with the method of  the normal deconvolution layer and nearest neighbor interpolation to increase the resolution, \cite{shi2016deconvolution, yang2018deep} further indicates that this method of increasing resolution by rearranging points in multiple LR output feature channels has good performance.
Therefore, considering the structure similarity between optical flow data and image data structure, we apply this idea in the process of HR optical flow field reconstruction.
Specifically,
in the final phase of the network, for $ F_{R} $ obtianed in the residual learning network, we use three convolutional layers to control the output feature maps, which can be represented as

\begin{equation}
F_{out} = C_{con}(F_{R})
\label{eq4} 
\end{equation}

Where $ F_{out} $ represents the output feature-maps of $ F_{R} $ after three convolutions $ C_{con} $.
Then we utilize the PS operation in ESPCN \cite{shi2016real} as a more efficient interpolation to get HR output.
Finally a $ 1\times1 $ convolution operation is used to obtain the final high resolution optical flow output of the network.
The final output of the network is expressed as

\begin{equation}
HR =C_{up}(F_{out})
\label{eq5} 
\end{equation}

Where $ HR $ represents the HR optical flow field of the network's final output, $ C_{up} $ represents the composite operation of PS and convolution.

\section{Training Data}

Unlike the image SR task, it is extremely difficult to obtain the ground turth of the optical flow from the real image sequence \cite{horn1981determining}.
Almost all optical flow estimations using CNN are primarily trained on synthetic data sets\cite{Dosovitskiy_2015_ICCV,ranjan2017optical,zhu2017guided,ilg2017flownet}, which may result in over-fitting on synthetic datasets but do not perform well on real data.
Flying Chairs\cite{Dosovitskiy_2015_ICCV}, FlyingThings3D\cite{mayer2016large}(Things3D), KITTI \cite{doi:10.1177/0278364913491297} and  MPI Sintel\cite{butler2012naturalistic} datasets are most commonly used datasets for optical flow estimation, where the KITTI dataset and MPI Sintel dataset are currently the most challenging and widely used optical flow benchmarks.

FlyingChairs\cite{Dosovitskiy_2015_ICCV} is obtained by applying affine transformation to the publicly available rendering atlas of images collected from Flickr and 3D chair models. This is a synthetic dataset containing $ 22872 $ image pairs and optical flow fields, providing enough data for training a CNN.
The MPI Sintel\cite{butler2012naturalistic} dataset is derived from the open source 3D animated short film Sintel and comes in two versions: clean and final.
Each version contains 1041 pair of images and a true value of the dense optical stream. The difference between the two is that the final version contains motion blur, camera noise and strong atmospheric effects such as fog. These effects are not included in the Clean version.
\paragraph{Training Data for Optical Flow Super-Resolution}
As mentioned in the part of network structure, the input data of our network are the LR optical flow field calculated by the existing method and the corresponding LR image in the first frame.
In the process of network training, we mainly choose FlowNetS and PWC-net as existing methods of optical flow estimation, both of which are outstanding in using convolutional neural network to learn optical flow. The two methods differ in their performance on different datasets. The FlowNetS outperform on the FlyingChairs Dataset, while PWC-net show better performance on
 the MPI Sintel Dataset.
In this paper, we use these two optical flow estimation methods to calculate the LR optical flow training data with $ 1/2 $ image resolution on the FlyingChairs and Sintel datasets separately.
In order to obtian the same resolution of the initial optical flow , the first frame images used for calculating the optical flows in these two datasets are downsampled to $ 1/2 $ resolution by bicubic interpolation.

In the following paragraphs, we still use FlyingChairs and Sintel to represent the LR optical flow data calculated by one of the methods above.
For each method, there are 22,872 LR optical flow fields with a resolution of $ 128\times96 $ in the FlyingChairs dataset, and 1,041 LR optical flow data with a resolution of $ 256\times109 $ from Sintel training set.
In order to observe overfitting of the network during training, following \cite{Dosovitskiy_2015_ICCV} the FlyingChairs were divided into 22,232 training and 640 test samples.
And we split the Sintel training set into 937 training and 104 test samples.

\section{Experimental Results}
\subsection{Network and Training Details}

In our proposed optical flow estimation network,
The low resolution optical flow field and the first image for optical flow estimation are concatenated feeding into the network.
The input low resolution optical flow field is calculated by the existing optical flow estimation method, the FlowNetS and the PWC-Net are used to calculate input data for training and test. 

In the feature extraction network, the convolution filter size of the $ Ci (i = 1, 2,3) $ convolution operation is $ 7\times7 $, $ 5\times5 $, $3\times3 $ respectively.
The convolution filter size of the $ C_{FE} $ operation for further feature extraction is $ 3\times3 $, and the activation function ReLU\cite{glorot2011deep} is used after each layer convolution.
In the residual learning network, we use the Residual dense block proposed in \cite{zhang2018residual} to fully learn the hierarchical features of each convolutional layer. After convergence analysis with different values of parameters on the part, there are $ 3 $ convolution layers in each block, $ 64 $ filters in each layer and $ 5 $ blocks in total. All the convolution filter size of this part in the network are set to $ 3\times3 $.
In the third part of the network, three convolutional layers with filter size of $ 5\times5 $, $ 3\times3 $, $3\times3 $ respectively futher extract feature maps and control feature map channels. The optical flow feature maps in LR space are then upscaled by the PS operation in \cite{shi2016real} followed by a $1\times1 $  Conv layer, and the network finally outputs high-resolution $ x-y $ optical flow of two channels.
For all convolution layers of the entire network, the stride is set to 1, and the zero padding keeps the size of the feature maps unchanged.

As training loss, we use endpoint error\cite{fortun2015optical}(EPE), which is the commonly used standard error measures for optical flow estimation. defined as the Euclidean distance between the HR flow vector and the ground truth.  
We training our network with a Tensorflow\cite{abadi2016tensorflow} framwork and choose Adam\cite{kingma2014adam} to update it. We use a small input batch of $ 4 $ LR image and flow pairs and initially set the learning rate to $ 10^{-4} $ for all layers, after the first 100 epochs, it was then divided by 2 every 50 epochs.

A large training set contains different object types and montion displacement is really necessary for training a network to get accurate HR optical flow field, we first training our network on FlyingChairs taining set calculated by FlowNetS and PWC-Net, then we fine-tune on the final version of the Sintel training set with a low learning rate of $ 10^{-6} $ for several hundreds epochs since the KITTI dataset only include sparse flow g                                                                                                                                                                                                                                                                                                                                                                    round truth.

\subsection{Results}

\begin{table}
\centering
 \begin{tabular}{c|c|c c|c c}%一个c表示有一列，格式为居中显示(center)
 \hline
 
 Method & FlyingChairs & \multicolumn{2}{|c|}{Sintel Clean} & \multicolumn{2}{|c}{Sintel Final}\\
     & test & train & test & train & test\\
 \hline % 在表格最下方绘制横线
 FlowNetS & 2.196 & 5.138 & 4.699 & 6.326 & 6.033\\%第二行第一列和第二列  中间用&连接
 %\hline % 在表格最下方绘制横线s
 Ours & 1.878 & 4.395 & 4.362 & 4.761 & 5.055\\
 \hline
 PWC-Net & 2.499 & 2.112 & 1.968 & 2.737 & 2.849\\
 
 Ours & 2.047 & 1.888 & 1.788 & 2.047 & 2.554\\
 \hline
 
 \end{tabular}
 \caption{Average endpoint errors (in pixels) of our method based on two well-performing methods compared to these methods using bilinear upsample on different datasets with scaling factor $ \times2 $.}
 \label{tab2} 
\end{table}

Table \ref{tab2} shows the endpoint error (EPE) of our network with the input LR optical flow predicted by two well-performing methods, FlowNetS and PWC-net, compared to these methods on public datasets(FlyingChairs, MPI-Sintel).Our network based on these two methods outperform the original results of both methods.

 \begin{figure}[htb]
 \centering
  \begin{subfigure}[h]{.19\linewidth}
    \centering
   Ground truth
   
    \includegraphics[width=.99\textwidth]{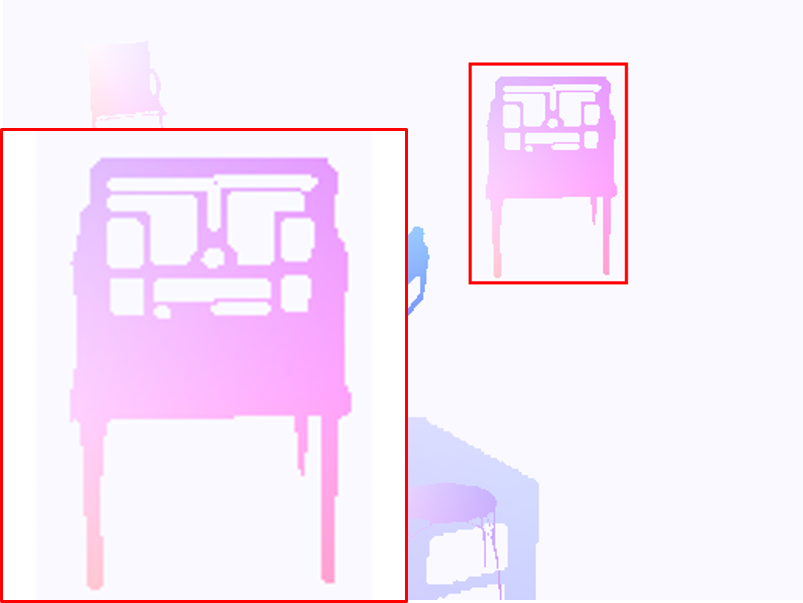}
    
  \end{subfigure}%   
  \begin{subfigure}[h]{.19\linewidth}
    \centering
     FlowNetS
     
    \includegraphics[width=.99\textwidth]{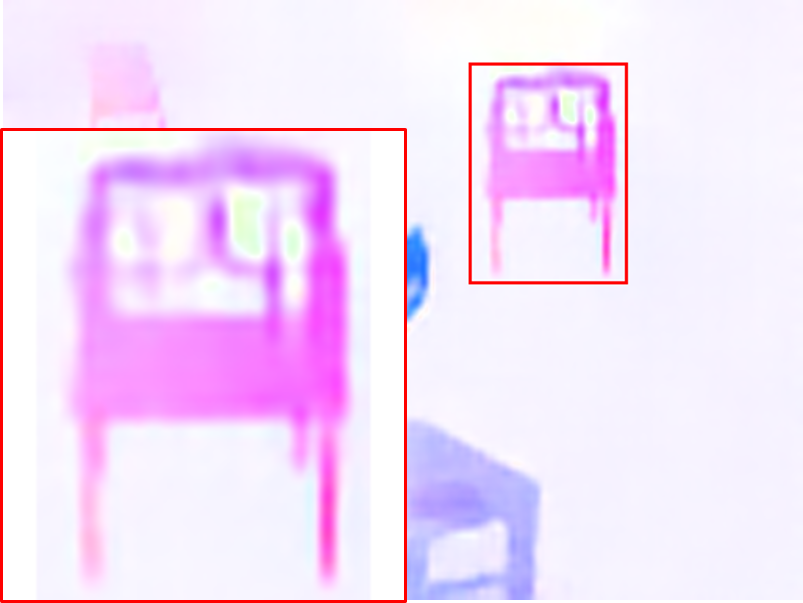}

  \end{subfigure}%  
  \begin{subfigure}[h]{.19\linewidth}
    \centering
    Ours
    
    \includegraphics[width=.99\textwidth]{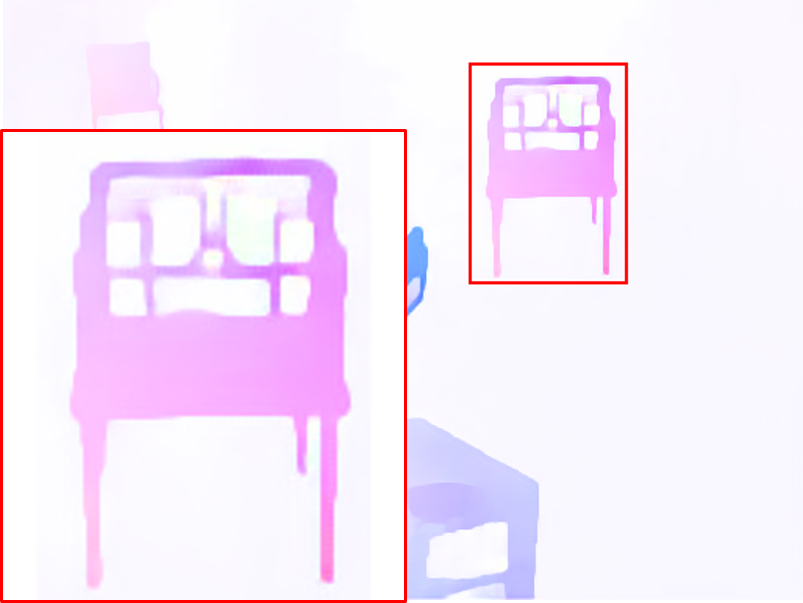}
    
    \end{subfigure}%\\ 
  \begin{subfigure}[h]{.19\linewidth}
    \centering
    PWC-Net
    
    \includegraphics[width=.99\textwidth]{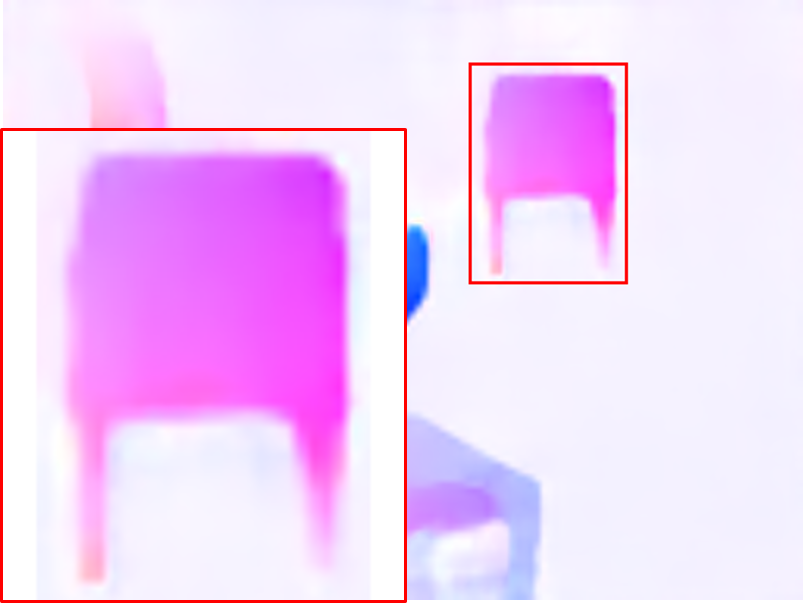}
    
  \end{subfigure}%\\ 
  \begin{subfigure}[h]{.19\linewidth}
    \centering
    Ours
    
    \includegraphics[width=.99\textwidth]{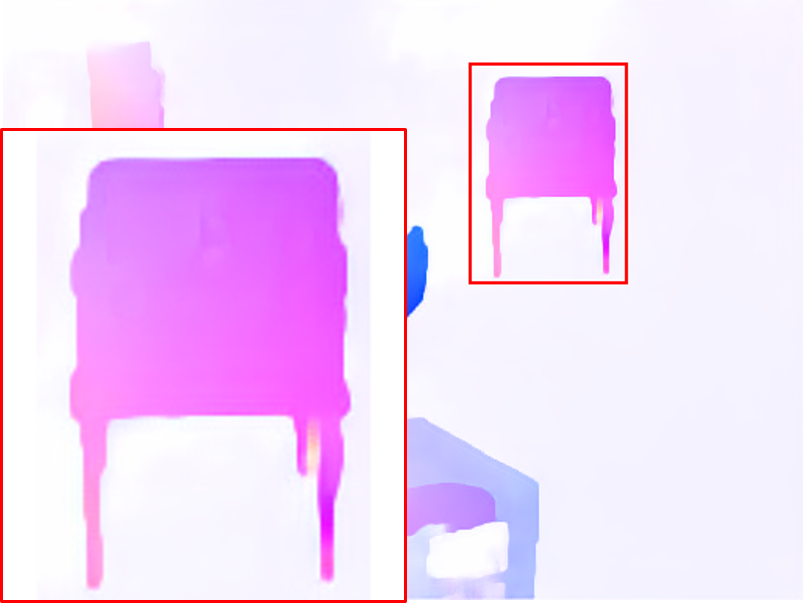}\\
    
  \end{subfigure}%  

  \begin{subfigure}[h]{.19\linewidth}
    \centering
    \includegraphics[width=.99\textwidth]{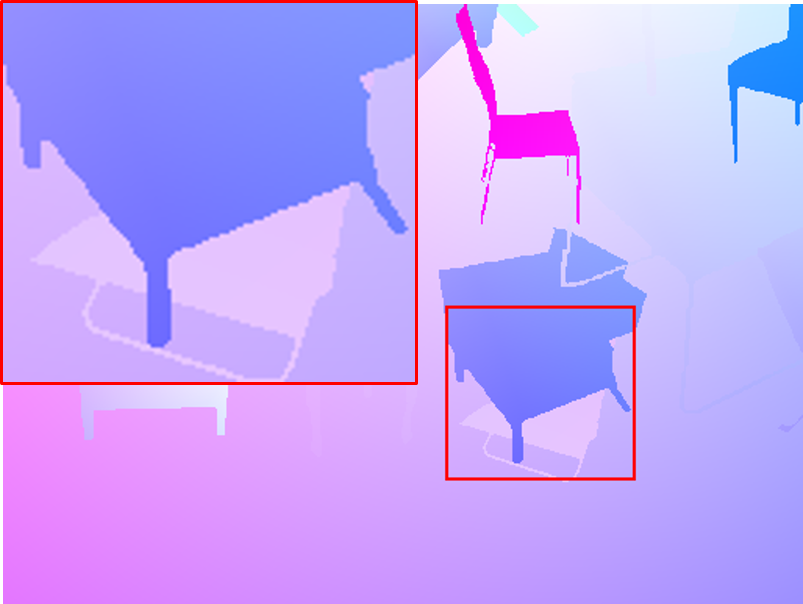}
  \end{subfigure}%   
  \begin{subfigure}[h]{.19\linewidth}
    \centering
    \includegraphics[width=.99\textwidth]{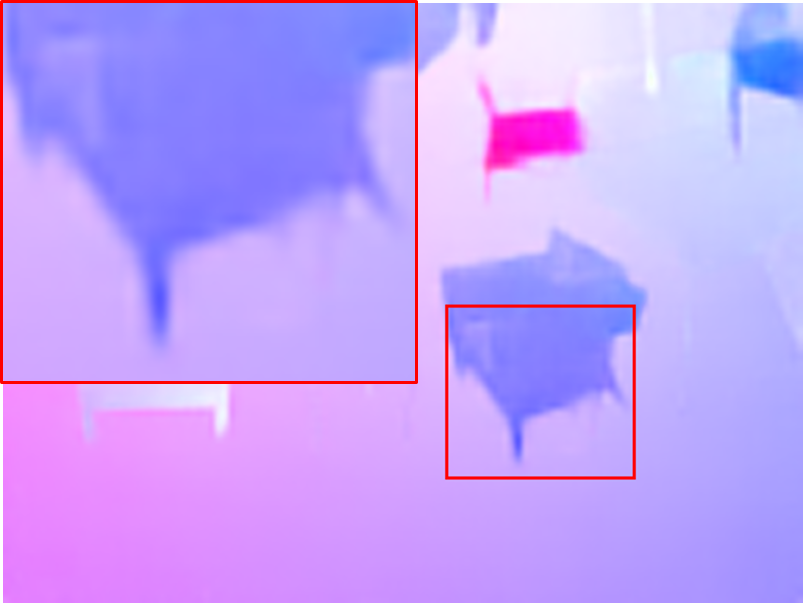}
  \end{subfigure}%  
  \begin{subfigure}[h]{.19\linewidth}
    \centering
    \includegraphics[width=.99\textwidth]{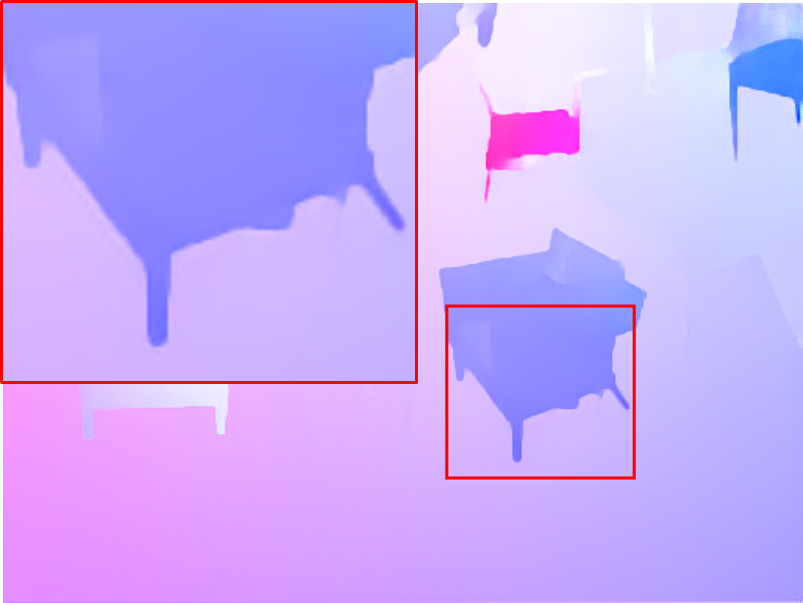}
  \end{subfigure}%\\
  \begin{subfigure}[h]{.19\linewidth}
    \centering
    \includegraphics[width=.99\textwidth]{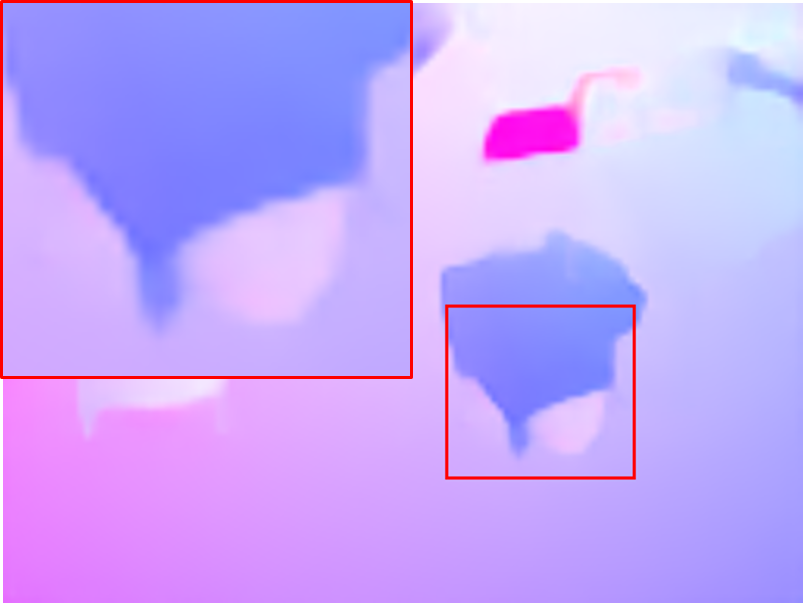}
  \end{subfigure}%  
  \begin{subfigure}[h]{.19\linewidth}
    \centering
    \includegraphics[width=.99\textwidth]{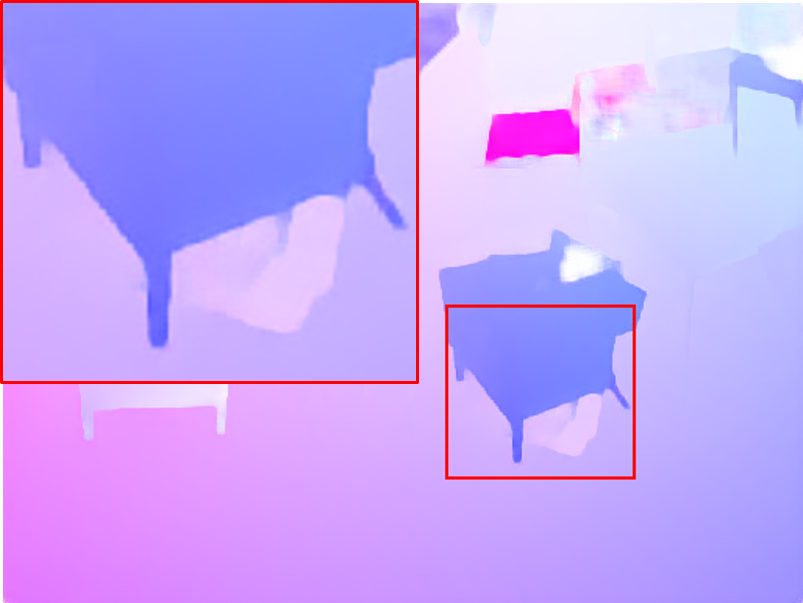}
  \end{subfigure}%\\

     \begin{subfigure}[h]{.19\linewidth}
    \centering
    \includegraphics[width=.99\textwidth]{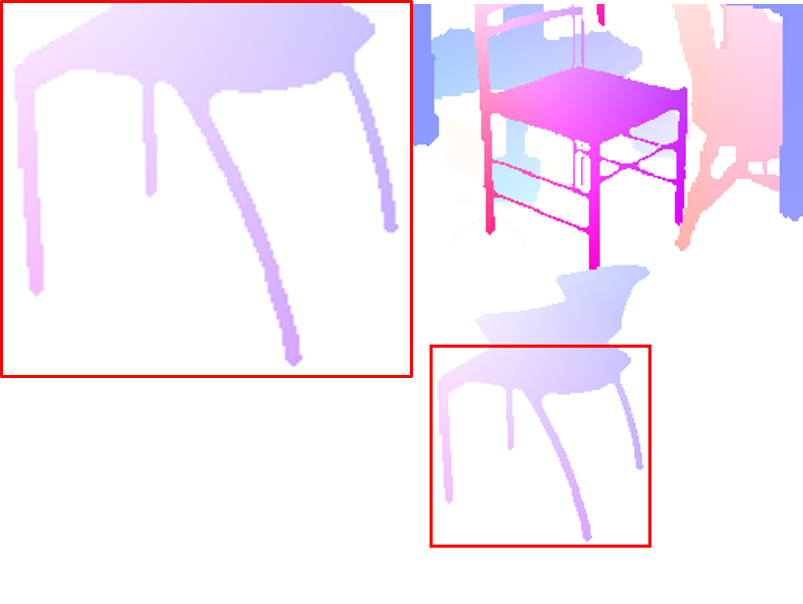}
  \end{subfigure}%  
  \begin{subfigure}[h]{.19\linewidth}
    \centering
    \includegraphics[width=.99\textwidth]{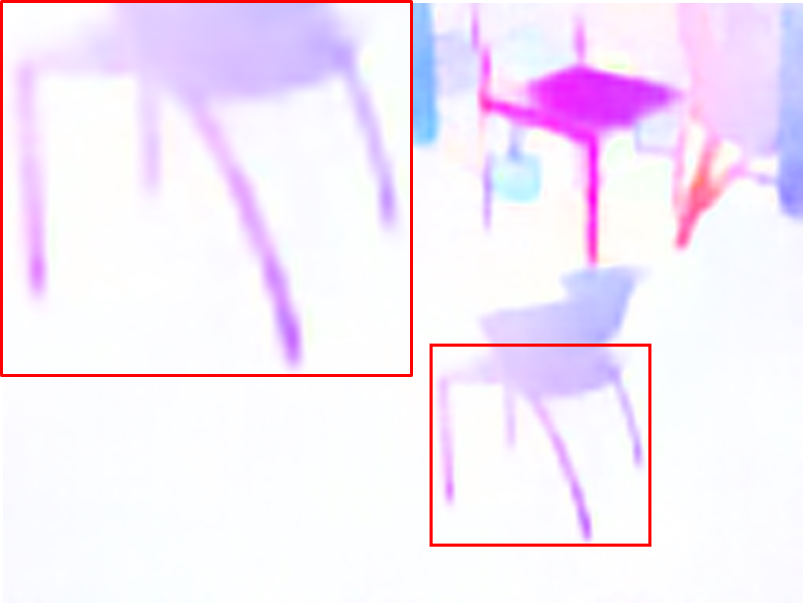}
  \end{subfigure}%
  \begin{subfigure}[h]{.19\linewidth}
  \centering
      \includegraphics[width=.99\textwidth]{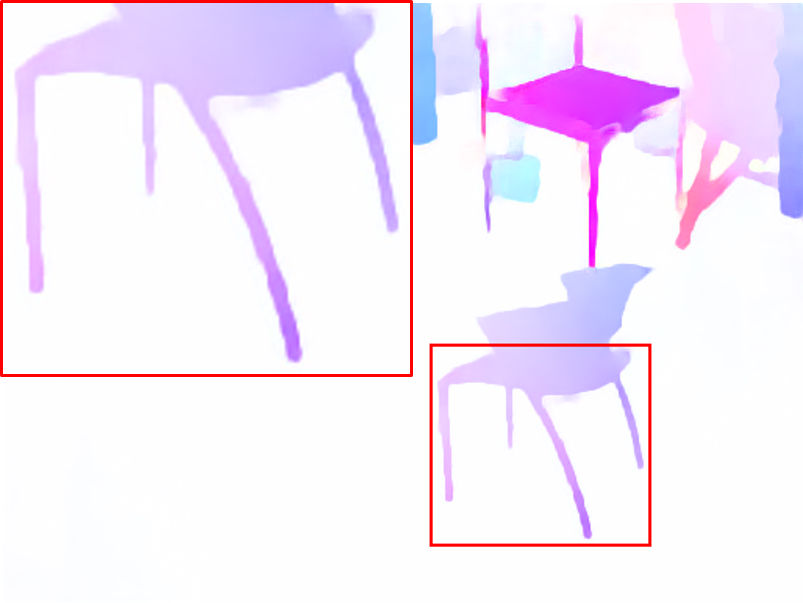}
  \end{subfigure}%  
  \begin{subfigure}[h]{.19\linewidth}
    \centering
    \includegraphics[width=.99\textwidth]{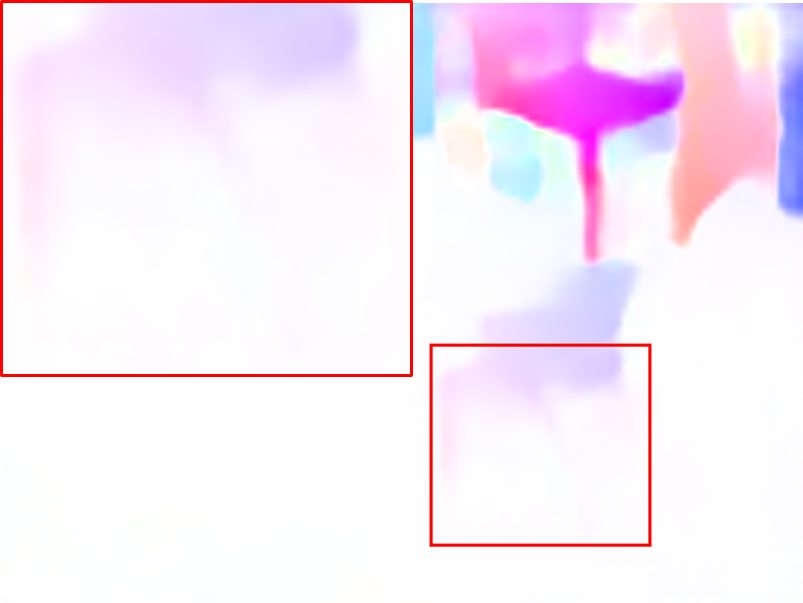}
  \end{subfigure}%\\
  \begin{subfigure}[h]{.19\linewidth}
    \centering
    \includegraphics[width=.99\textwidth]{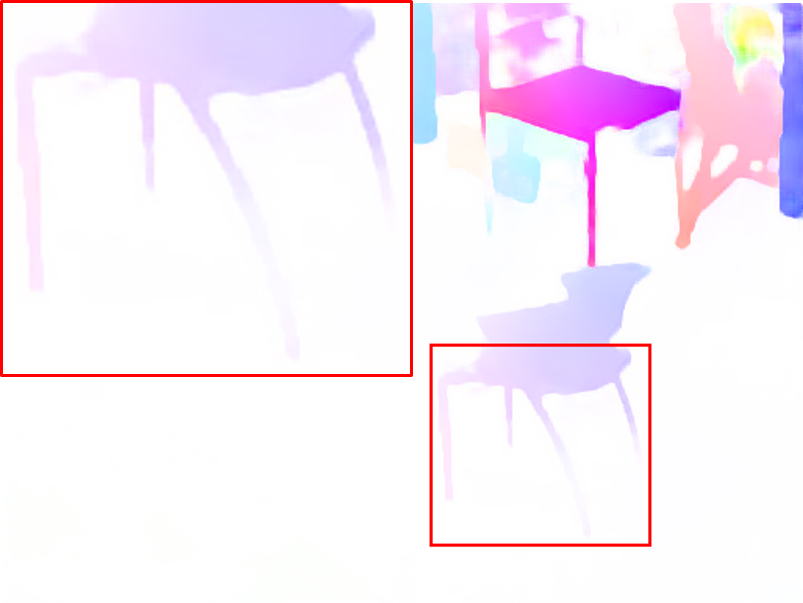}
  \end{subfigure}%\\

  \caption{Optical flow super-resolution examples from the FlyingChairs dataset compared to FlowNetS and PWC-Net.}\label{fig:4}
\end{figure}

\begin{figure}[htb]
\centering
  \begin{subfigure}[h]{.16\linewidth}
    \centering
    \begin{tiny}
   Images
   \end{tiny}   
    \includegraphics[width=.99\textwidth]{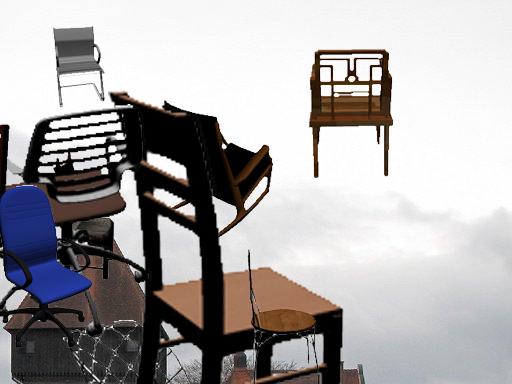}
  \end{subfigure}%   
  \begin{subfigure}[h]{.16\linewidth}
    \centering
     \begin{tiny}
      Ground truth
     \end{tiny}
    \includegraphics[width=.99\textwidth]{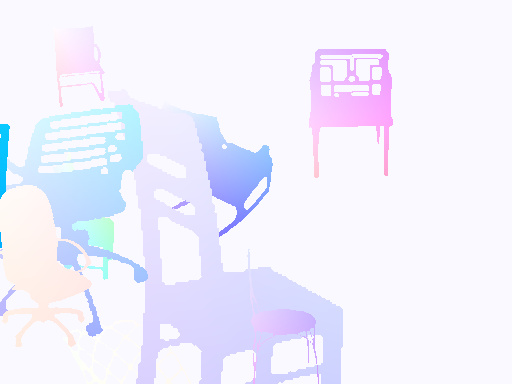}
  \end{subfigure}%  
  \begin{subfigure}[h]{.16\linewidth}
    \centering
    \begin{tiny}
    FlowNetS
    \end{tiny}
    \includegraphics[width=.99\textwidth]{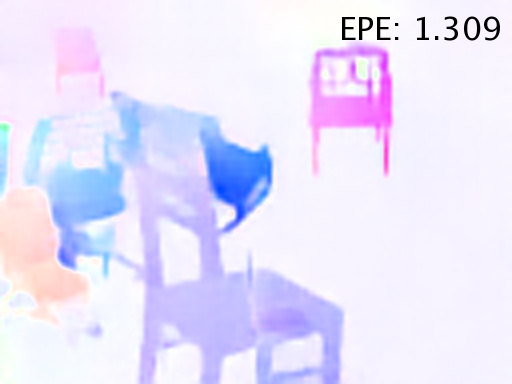}
  \end{subfigure}%  
  \begin{subfigure}[h]{.16\linewidth}
    \centering
    \begin{tiny}
     Ours
    \end{tiny}
    \includegraphics[width=.99\textwidth]{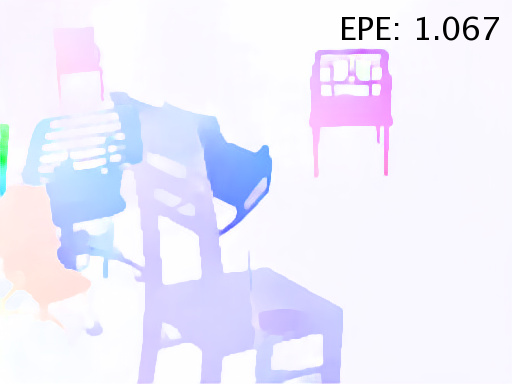}
  \end{subfigure}%   
  \begin{subfigure}[h]{.16\linewidth}
    \centering
    \begin{tiny}
    PWC-net
    \end{tiny}
    \includegraphics[width=.99\textwidth]{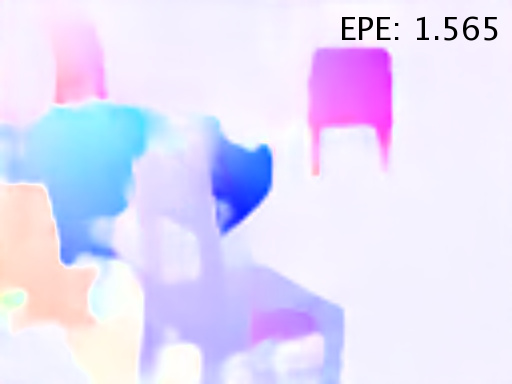}
  \end{subfigure}%
  \begin{subfigure}[h]{.16\linewidth}
    \centering
    \begin{tiny}
     Ours
    \end{tiny}
    \includegraphics[width=.99\textwidth]{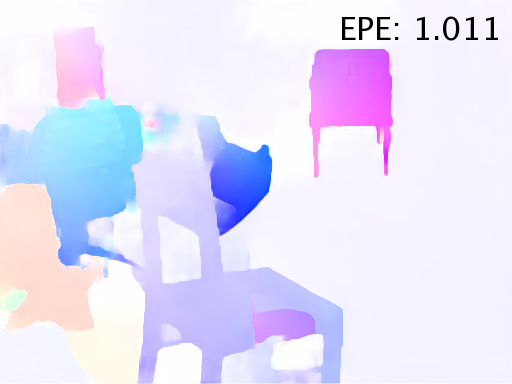}
  \end{subfigure}%

  \begin{subfigure}[h]{.16\linewidth}
    \centering
    \includegraphics[width=.99\textwidth]{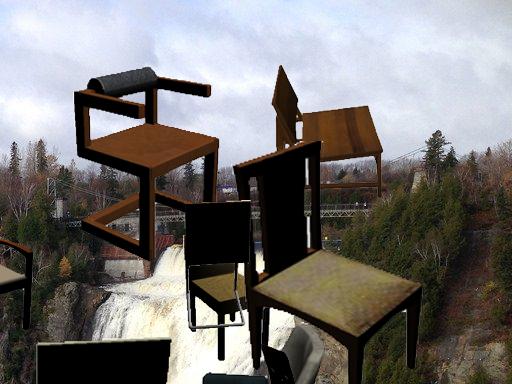}
  \end{subfigure}%   
  \begin{subfigure}[h]{.16\linewidth}
    \centering
    \includegraphics[width=.99\textwidth]{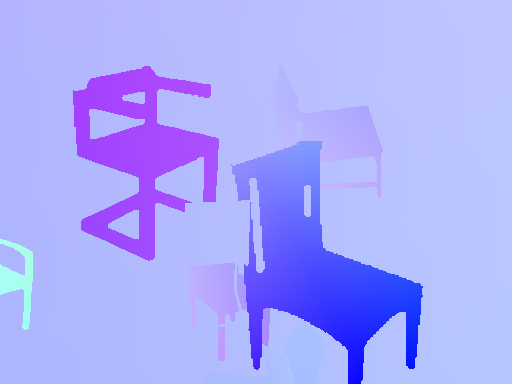}
  \end{subfigure}%  
  \begin{subfigure}[h]{.16\linewidth}
    \centering
    \includegraphics[width=.99\textwidth]{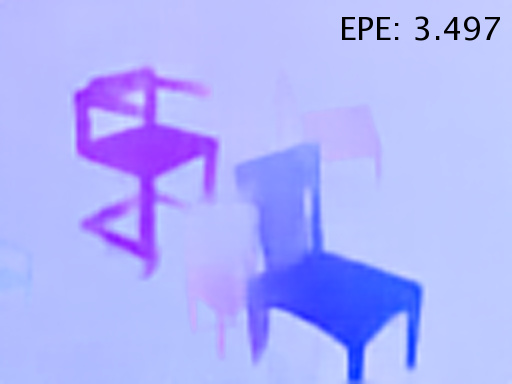}
  \end{subfigure}%  
  \begin{subfigure}[h]{.16\linewidth}
    \centering
    \includegraphics[width=.99\textwidth]{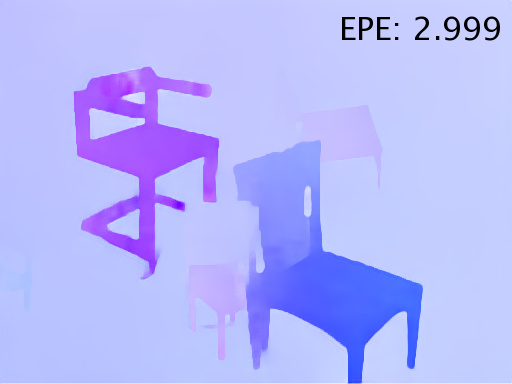}
  \end{subfigure}%   
  \begin{subfigure}[h]{.16\linewidth}
    \centering
    \includegraphics[width=.99\textwidth]{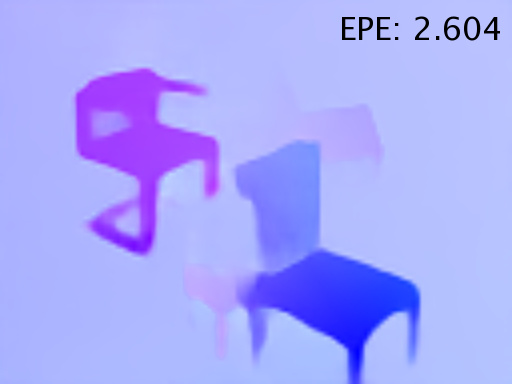}
  \end{subfigure}%
  \begin{subfigure}[h]{.16\linewidth}
    \centering
    \includegraphics[width=.99\textwidth]{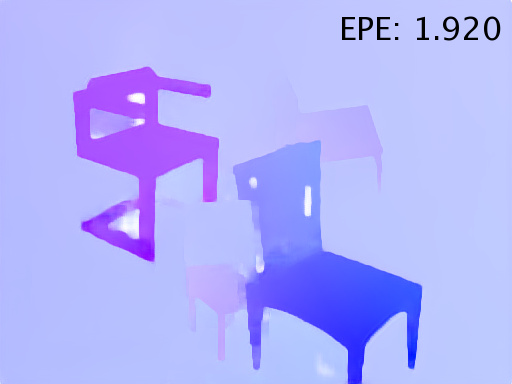}
  \end{subfigure}%

  \begin{subfigure}[h]{.16\linewidth}
    \centering
    \includegraphics[width=.99\textwidth]{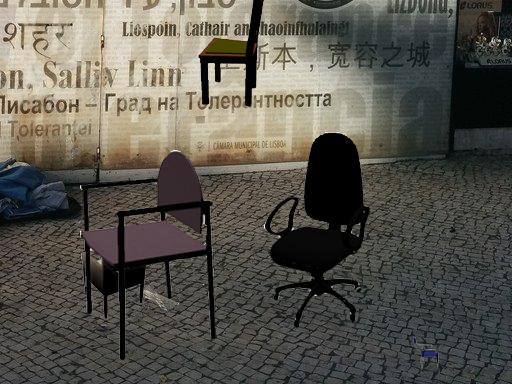}
  \end{subfigure}%   
  \begin{subfigure}[h]{.16\linewidth}
    \centering
    \includegraphics[width=.99\textwidth]{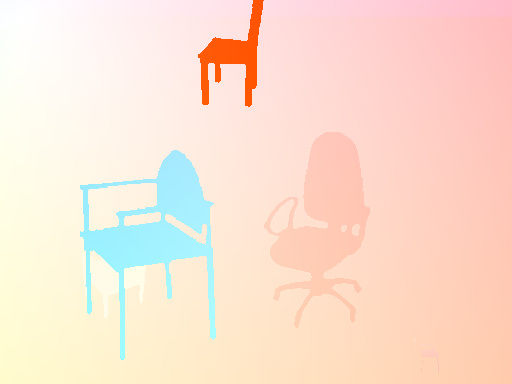}
  \end{subfigure}%  
  \begin{subfigure}[h]{.16\linewidth}
    \centering
    \includegraphics[width=.99\textwidth]{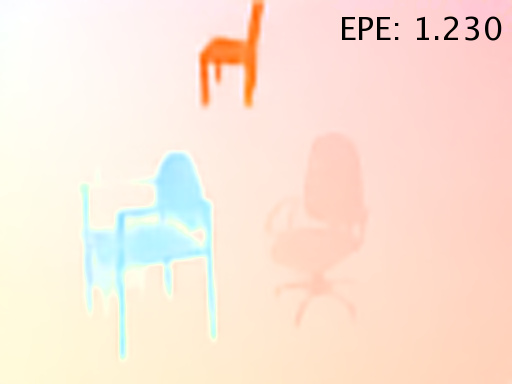}
  \end{subfigure}%  
  \begin{subfigure}[h]{.16\linewidth}
    \centering
    \includegraphics[width=.99\textwidth]{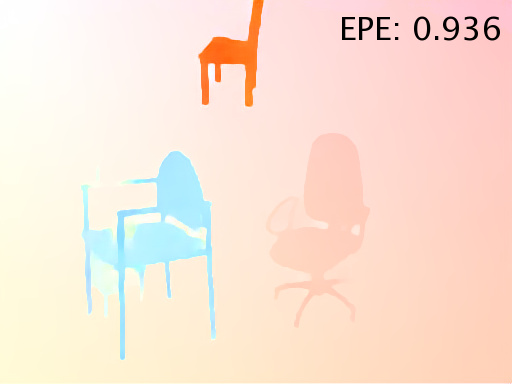}
  \end{subfigure}%   
  \begin{subfigure}[h]{.16\linewidth}
    \centering
    \includegraphics[width=.99\textwidth]{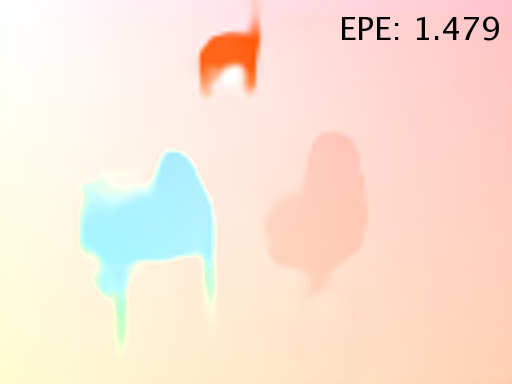}
  \end{subfigure}%
  \begin{subfigure}[h]{.16\linewidth}
    \centering
    \includegraphics[width=.99\textwidth]{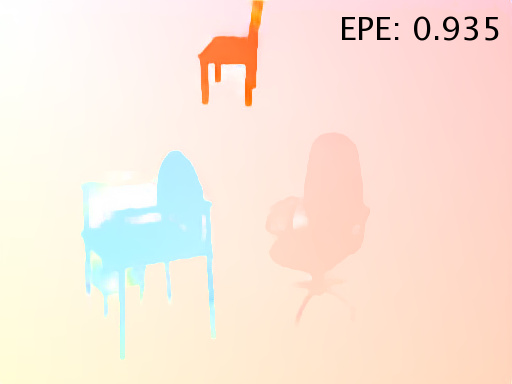}
  \end{subfigure}%

  \begin{subfigure}[h]{.16\linewidth}
    \centering
    \includegraphics[width=.99\textwidth]{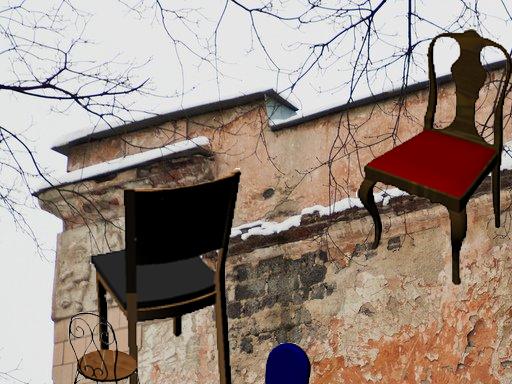}
  \end{subfigure}%   
  \begin{subfigure}[h]{.16\linewidth}
    \centering
    \includegraphics[width=.99\textwidth]{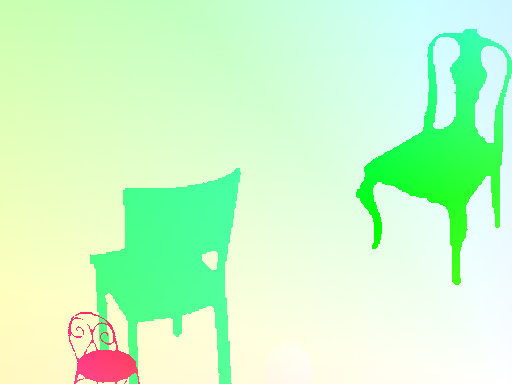}
  \end{subfigure}%  
  \begin{subfigure}[h]{.16\linewidth}
    \centering
    \includegraphics[width=.99\textwidth]{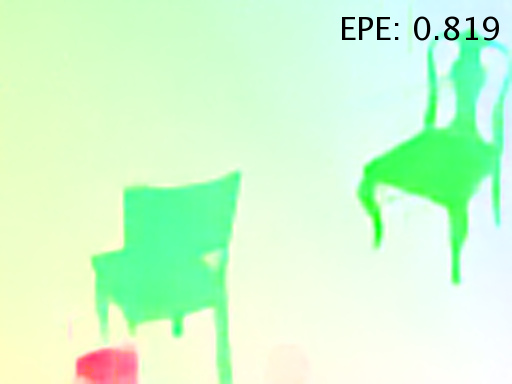}
  \end{subfigure}%  
  \begin{subfigure}[h]{.16\linewidth}
    \centering
    \includegraphics[width=.99\textwidth]{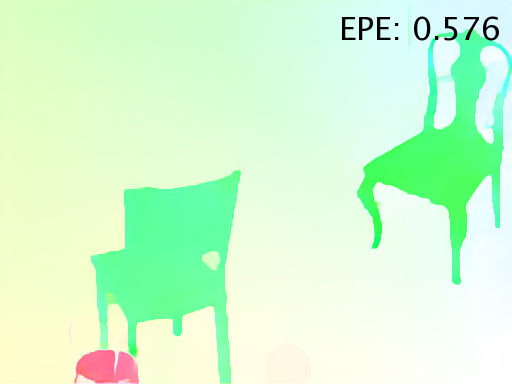}
  \end{subfigure}%   
  \begin{subfigure}[h]{.16\linewidth}
    \centering
    \includegraphics[width=.99\textwidth]{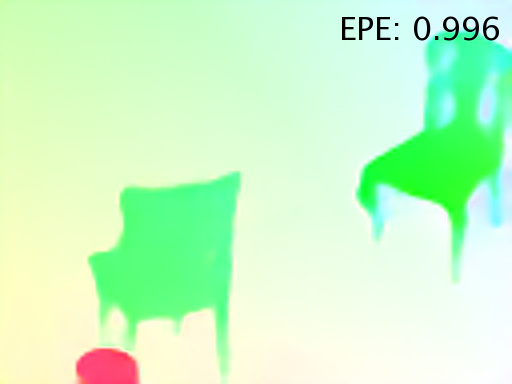}
  \end{subfigure}%
  \begin{subfigure}[h]{.16\linewidth}
    \centering
    \includegraphics[width=.99\textwidth]{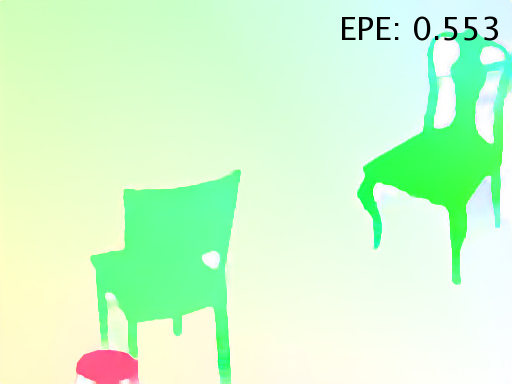}
  \end{subfigure}%

  \begin{subfigure}[h]{.16\linewidth}
    \centering
    \includegraphics[width=.99\textwidth]{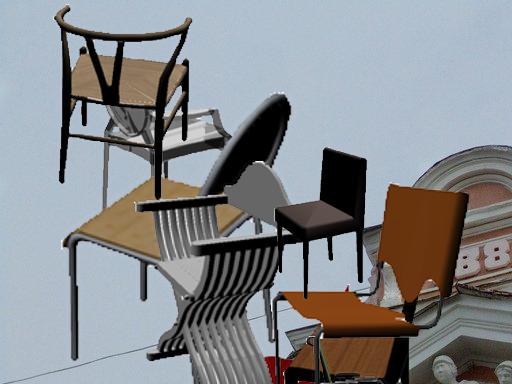}
  \end{subfigure}%   
  \begin{subfigure}[h]{.16\linewidth}
    \centering
    \includegraphics[width=.99\textwidth]{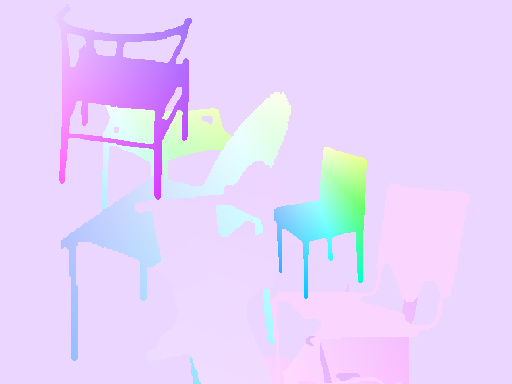}
  \end{subfigure}%  
  \begin{subfigure}[h]{.16\linewidth}
    \centering
    \includegraphics[width=.99\textwidth]{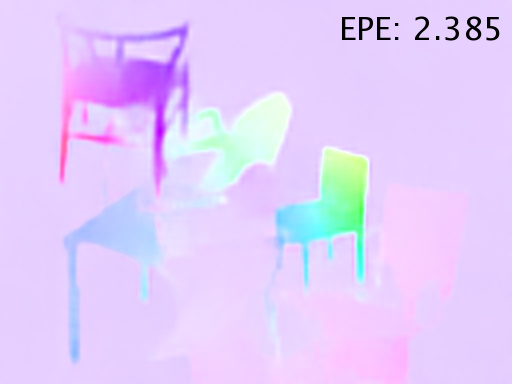}
  \end{subfigure}%  
  \begin{subfigure}[h]{.16\linewidth}
    \centering
    \includegraphics[width=.99\textwidth]{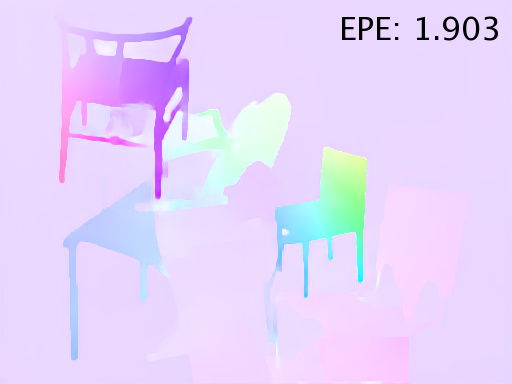}
  \end{subfigure}%   
  \begin{subfigure}[h]{.16\linewidth}
    \centering
    \includegraphics[width=.99\textwidth]{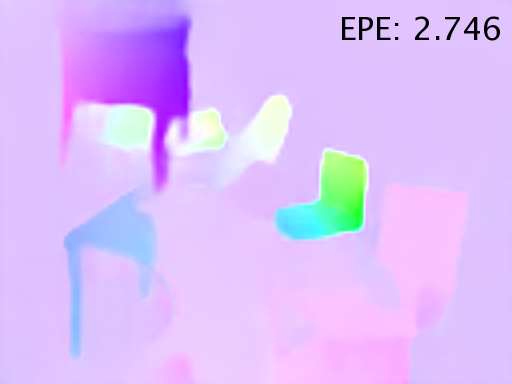}
  \end{subfigure}%
  \begin{subfigure}[h]{.16\linewidth}
    \centering
    \includegraphics[width=.99\textwidth]{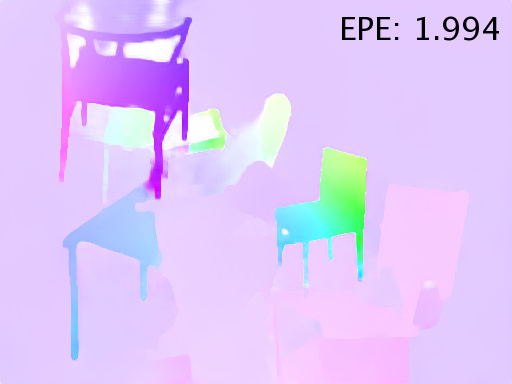}
  \end{subfigure}%

  \begin{subfigure}[h]{.16\linewidth}
    \centering
    \includegraphics[width=.99\textwidth]{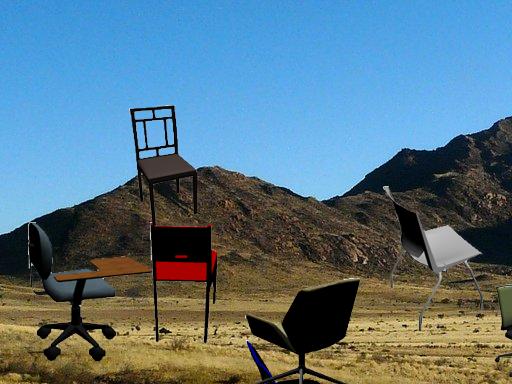}
  \end{subfigure}%   
  \begin{subfigure}[h]{.16\linewidth}
    \centering
    \includegraphics[width=.99\textwidth]{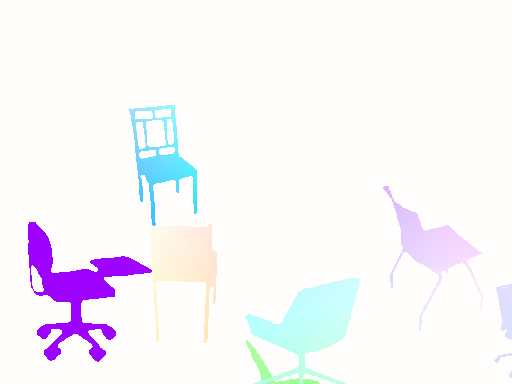}
  \end{subfigure}%  
  \begin{subfigure}[h]{.16\linewidth}
    \centering
    \includegraphics[width=.99\textwidth]{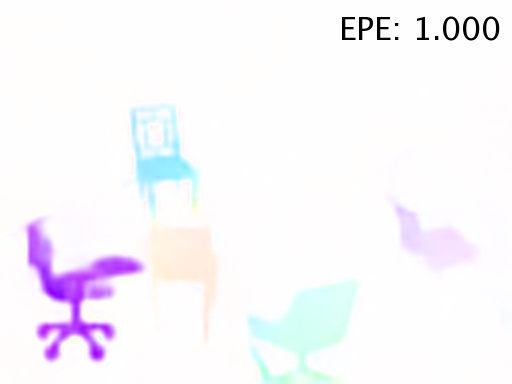}
  \end{subfigure}%  
  \begin{subfigure}[h]{.16\linewidth}
    \centering
    \includegraphics[width=.99\textwidth]{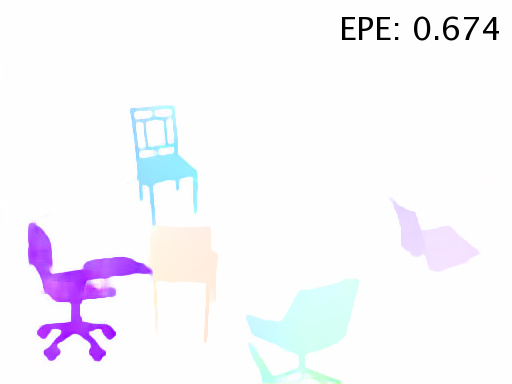}
  \end{subfigure}%   
  \begin{subfigure}[h]{.16\linewidth}
    \centering
    \includegraphics[width=.99\textwidth]{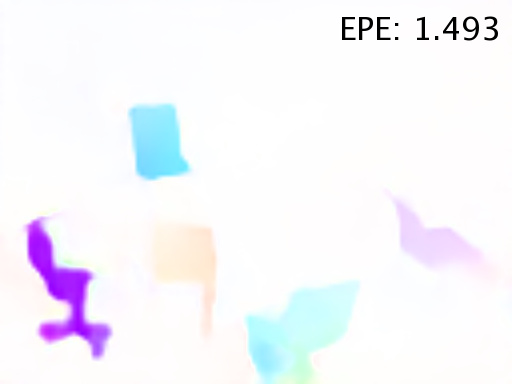}
  \end{subfigure}%
  \begin{subfigure}[h]{.16\linewidth}
    \centering
    \includegraphics[width=.99\textwidth]{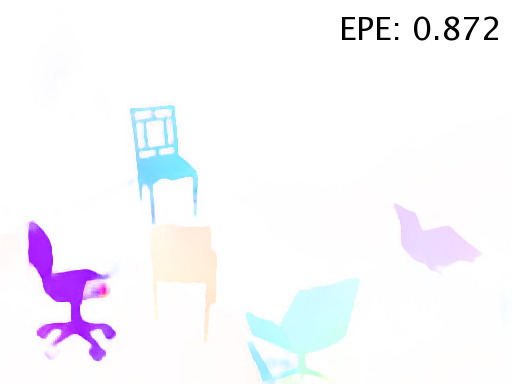}
  \end{subfigure}%
     
  \caption{ Examples of optical flow SR on the FlyingChairs dataset with scaling
factor $ \times2 $. In each row left to right: first frame image, ground truth flow, FlowNetS:  bilinear upsample, Our SR result and PWC-net: bilinear upsample, Our SR result. Endpoint error is shown for every frame.}\label{fig:1}
\end{figure}

The color code visualization method allows for dense visualization of the flow field and better visual perception of subtle differences between adjacent motion vectors\cite{fortun2015optical}. 
Some color code visualization examples on FlyingChairs with scaling
factor $ \times2 $ are shown in figure \ref{fig:4} and figure \ref{fig:1}. 
We can see that the FlowNetS perform better on synthetic FlyingChairs dataset than the PWC-Net, which outperform the FlowNetS method on Sintel. For our method based on FlowNetS, our 
network further suppresses the blurring artifacts and output more subpixel-accurate HR optical flow with sharper edges.
For our method based on the PWC-Net, as it is shown in figure \ref{fig:1}, our network significantly improve the result compared to the PWC-Net using bilinear upsample on FlyingChairs.
This comparison suggests that training on initial LR flow data predicted by two kinds of methods with different performance our method can not only obtain full image resolution but futher improve the accuracy of input optical flow.

It also indicates that using bilinear interpolation to restore the original resolution would introduce noise and blur optical flow boundaries and our method has strong ability to overcome that.
We can find that, compared with the original method test result on FlyingChairs, when applying our method to PWC-Net the performance improvement is greater than that when applying our method to FlowNetS. A similar conclusion can be drawn on the Sintel dataset, that is, when applying our method to FlowNet, we can obtain greater performance improvement. This suggests that our method can further improve the performance of the original method when applied to a certain optical flow estimation method, and there is more room for improvement of the general performance methods.

\begin{figure}[htbp]
\centering
\includegraphics[width=\textwidth]{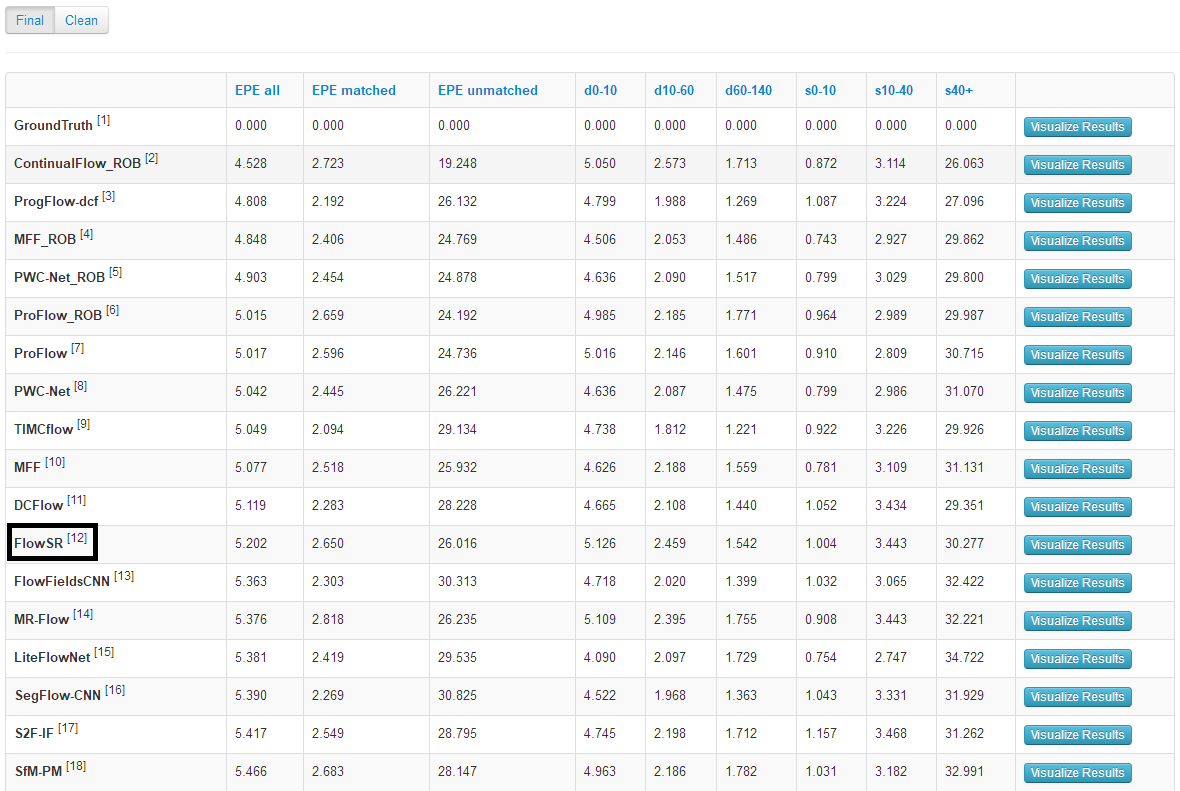}
\caption{The screen shots of the MPI Sintel final pass on August 28th, 2018}
\label{result}
\end{figure}

In order to apply our method to the actual method, we use the public PyTorch implementation of PWC-Net to predicte LR optical flow data with $ 1/2 $ image resolution on test set of MPI-Sintel Dataset, and apply our network to generate the full resolution optical flow field without other refinement operations. We name this method as FlowSR and the screen shots of the MPI Sintel final pass is shown in figure \ref{result}. Note that according to their public file, the PyTorch implementation is inferior to the Caffe implementation ($ ~3\% $ performance drop on Sintel). These are due to differences in implementation between Caffe and PyTorch, such as image resizing and I/O.

\subsection{Analysis}
In terms of the same data set, we used two methods with different performance on this data set to calculate LR optical flow respectively, and combined these two data sets into a new data set to train the network. As a result, the performance of both methods has been improved, and the performance of the general methods has been improved even more. This shows that the network has learned from the better optical flow estimation results to obtain the effective optical flow information from the first frame image to improve the poor results, and achieved the effect of drawing on the strengths and making up for the weaknesses.

Compare to the SISR problems: on the one hand, similar to SISR, the purpose of our OFSR task is to generate HR optical flow field with clearer boundary, more details and higher accuracy.
On the other hand, in this paper, our network is used to carry out super-resolution reconstruction of the low-resolution optical flow field obtained by using the existing optical flow estimation methods. The input training data itself contains estimation bias, which would easily amplify the erroneous flows. This is very different from SISR, the LR image are generally a bicubic downsampled, blurred, and noisy version of HR image.
To solve this problem we introduce the first frame image as the guidance information, constrain the optical flow field boundary, and provide more abundant information of object texture to suppress noise and recover details.
Moreover, our CNN model is much smaller for the SISR model are usually perform better with much deeper CNN model. 
The experimental results show that the network output HR optical flow field with lower EPE, sharper edges than original results,
which shows that the first picture can play a role in optimizing the original optical flow results. To sum up, the purpose of our SR network is of more practical significance.

\section{Conclusions}
In this paper, we propose a optical flow SR convolution neural network
model based on image guidance for more accurate HR optical flow, where the
LR optical flow is calculated by some existing method and the image is the first
frame image used in optical flow estimation. Our model takes the first frame
image as a guidance to provide abundant graphic boundary information for
the optical flow field. Using various method of SISR for reference, we perform
feature extraction in LR space for residual learning. We apply our method
to two well-performing methods and achieve favorable performance agains the
original method, which indicate the network has good ability to suppress noise,
restore detail and sharp boundaries. 

\section*{References}

\bibliography{bibfile}

\end{document}